\documentclass[journal]{IEEEtran}
\usepackage[colorlinks,linkcolor=red]{hyperref}

\usepackage{amsmath}
\usepackage{epsfig}
\usepackage{graphicx}
\usepackage{subfigure}
\usepackage{multirow}
\usepackage{color}
\usepackage{changepage}
\usepackage{times}
\usepackage{float}
\usepackage{amssymb}
\usepackage{multirow}
\usepackage{bbding}
\usepackage{booktabs}
\usepackage{algorithm}
\usepackage{algpseudocode}
\usepackage{caption2}
\usepackage{verbatim}
\usepackage{dsfont}
\usepackage{amssymb}
\usepackage{stfloats}
\usepackage{tikz}
\usetikzlibrary{arrows}

\usepackage{url}
\usepackage{xurl}

\ifCLASSINFOpdf
\else
\fi

\begin{document}
%
\title{When Face Recognition Meets Occlusion: A New Benchmark}
%
%
%

\author{Baojin~Huang, Zhongyuan~Wang, Guangcheng Wang, Kui Jiang, Kangli Zeng, Zhen Han, Xin Tian, Yuhong Yang
\thanks{B. Huang, Z. Wang, G. Wang, K. Jiang, K. Zeng, H. Zhen, T. Xin, Y. Yang are with the National Engineering Research Center for Multimedia Software, School of Computer Science, Wuhan University, Wuhan 430072, China. Thanks to National Natural Science Foundation of China (U1903214, U1736206, 62071339, 62072347,61971315, 62072350) for funding. The numerical calculations in this paper have been done on the supercomputing system in the Supercomputing Center of Wuhan University. \emph{(Corresponding author: Zhongyuan Wang, wzy\_hope@163.com).}}
}

%
%

\maketitle

\begin{abstract}
	The existing face recognition datasets usually lack occlusion samples, which hinders the development of face recognition. Especially during the COVID-19 coronavirus epidemic, wearing a mask has become an effective means of preventing the virus spread. Traditional CNN-based face recognition models trained on existing datasets are almost ineffective for heavy occlusion. To this end, we pioneer a simulated occlusion face recognition dataset. In particular, we first collect a variety of glasses and masks as occlusion, and randomly combine the occlusion attributes (occlusion objects, textures,and colors) to achieve a large number of more realistic occlusion types. We then cover them in the proper position of the face image with the normal occlusion habit. Furthermore, we reasonably combine original normal face images and occluded face images to form our final dataset, termed as Webface-OCC. It covers 804,704 face images of 10,575 subjects, with diverse occlusion types to ensure its diversity and stability. Extensive experiments on public datasets show that the ArcFace retrained by our dataset significantly outperforms the state-of-the-arts. Webface-OCC 
	is available at \url{https://github.com/Baojin-Huang/Webface-OCC}.
	
\end{abstract}

\section{Introduction}
\label{sec:intro}

With the development of face recognition technologies based on deep learning, many face recognition methods \cite{schroff2015facenet,liu2017sphereface,deng2018arcface,liu2019fair} under normal scenes have achieved impressive performance, even exceeding the human recognition ability on the benchmark dataset \cite{huang2008labeled}. However, when the face image is occluded in the actual unrestricted scene, the recognition accuracy drops sharply. 
To eliminate the influence of occlusion on face recognition accuracy, researchers mostly train deep networks to be "familiar" with the occluded area of face images, thereby weakening or inpainting occlusion components. Note that there is currently no open-source occlusion face recognition dataset. The existing occluded face recognition methods based on deep learning all synthesize their own occluded face images to train the network.

\begin{figure}[!t]
	\small
	\centering
	\vspace{-0.07cm}
	\hspace{-0.1cm}
	\includegraphics[width=0.99\linewidth]{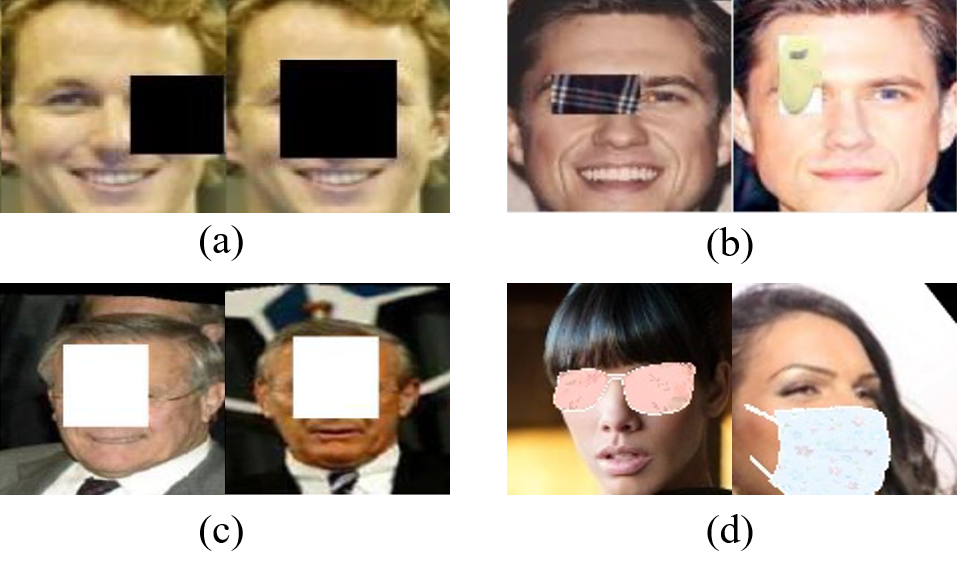}\\
	\vspace{-0.1cm}
	\caption{\small Some samples of training and testing images in recent occlusion face recognition papers. (a) MaskNet \cite{he2016masked}, (b) PSDN \cite{song2019occlusion}, (c) wID \cite{ge2020occluded}, (d) Ours.}
	\label{fig:01}
	\vspace{-0.3cm}
\end{figure}

\begin{figure*}[!t]
	\small
	\centering
	\vspace{0.07cm}
	\hspace{-0.1cm}
	\includegraphics[width=0.99\linewidth]{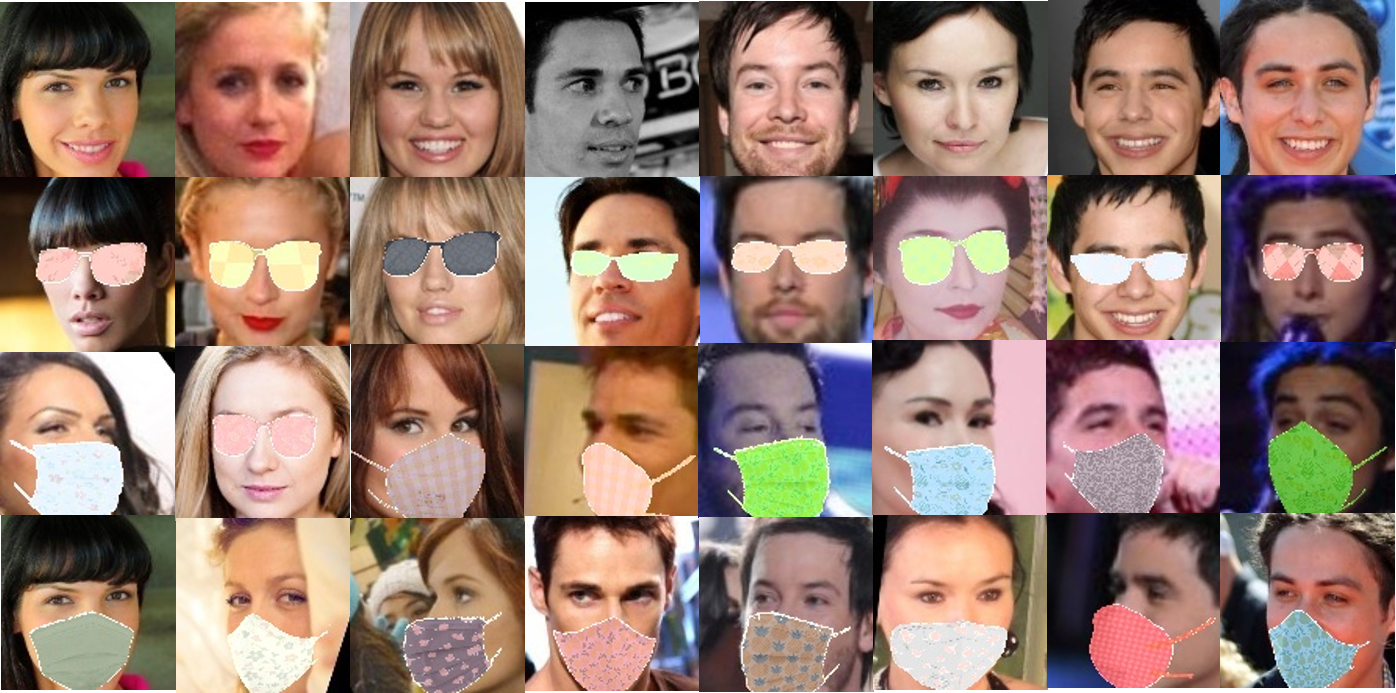}\\
	\vspace{-0.1cm}
	\caption{\small Samples of occlusion face images in Webface-OCC. The first row shows normal faces, and the second and third rows are their corresponding occlusion faces.}
	\label{fig:02}
	\vspace{-0.3cm}
\end{figure*}

As shown in Fig. \ref{fig:01}, we list some samples of occluded images in recent occlusion face recognition papers. Wan et al. \cite{he2016masked} synthesized occlusion images with random black block sizes n = 40, 50, 60, 70, respectively. Only a single occlusion type is unnatural and not robust. Song et al. \cite{song2019occlusion} masked face images with three types of occlusions. Although it increases the diversity of occlusion types, it is far from the actual occlusion situations (block position and size). Recently, to generate
occluded faces, Ge et al. \cite{ge2020occluded} take a m $\times$ m (e.g., m = 48) mask to cover on real face images randomly. Obviously, this synthesis method is too blunt to adapt to actual recognition needs. The GAN-based method \cite{9157070} can generate visually natural occluded face images, but objectively changes the image's detailed information. The recognizer trained by these images does not perform well in real scenes. In brief, to synthesize occlusion images, most of the current methods cover some abnormal occlusion masks to the normal image, deviating from the actual situation.

To address these drawbacks, we propose a Webface-OCC dataset to improve the performance of occluded face recognition in real scenes. In the Webface-OCC dataset, we fully consider the occlusion type and position and synthesize the occluded face image, as shown in Fig. \ref{fig:01} (d). In particular, we first collect a variety of occlusion types of glasses and masks, and randomly combine the occlusion attributes (occlusion objects, textures,and colors) to achieve a large number of more realistic occlusion types. We then cover them in the proper position of the face image with the normal occlusion habit. Besides, we design a reasonable combination method for simulating occlusion of face images. Extensive experimental results on simulated and real-world masked face datasets confirm our built dataset's substantial superiority in terms of accuracy without reducing the original effect on the general face recognition dataset.

\section{PROPOSED OCCLUSION FACE RECOGNITION DATASET}
\label{sec:format}
The CASIA-Webface \cite{yi2014learning} dataset, as a common face recognition dataset, contains a small part of occlusion samples. Various CNN-based face recognition models \cite{schroff2015facenet,liu2017sphereface,deng2018arcface,liu2019fair} trained on this dataset can achieve good performance in recognizing faces with small occlusions. It is conceivable that to improve the recognition performance of the model on occluded face images, a large-scale occlusion sample dataset is essential. To increase the proportion of occluded samples in the CASIA-Webface, we combine the key points of the face and multiple occlusion types to achieve occlusion simulation as much as possible authenticity and diversity. In this section, we will introduce the process of occlusion simulation, as well as show the basic statistics of our dataset.

\subsection{Occlusion Simulation}
\label{ssec:subhead}
Based on the fact that real face occlusion is to block the facial features of the face, we cover common occlusion types (glasses, masks, etc.) on the normal face image with the help of known key points of the face. We collect several occlusion templates from various natural scenes to enrich occlusion diversity and authenticity. Meanwhile, we adopt an accurate face key point detection model to obtain the face
image’s key point information.

\begin{figure}[!t]
	\small
	\centering
	\vspace{-0.07cm}
	\hspace{-0.1cm}
	\includegraphics[width=0.99\linewidth]{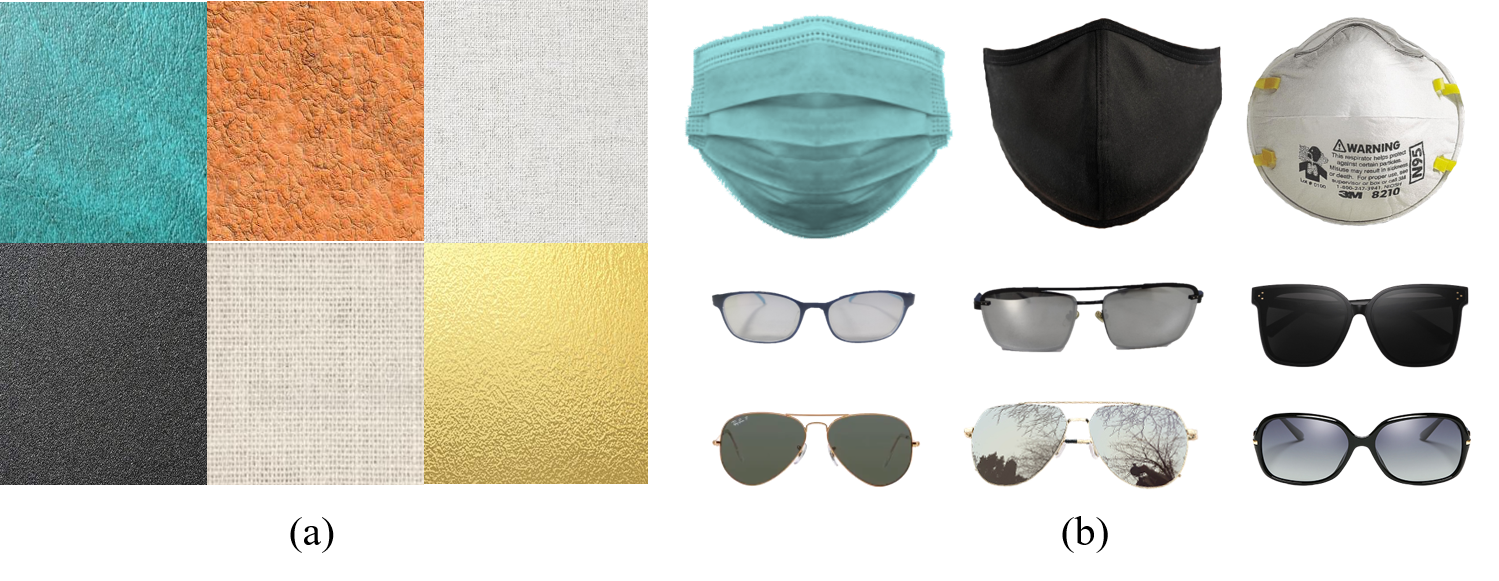}\\
	\vspace{-0.1cm}
	\caption{\small Texture of occluder (a) and type of occluder (b) in Fig. \ref{fig:02}.}
	\label{fig:03}
	\vspace{-0.3cm}
\end{figure}

For occlusion types, including occlusion objects, textures, and colors, as shown in Fig. \ref{fig:03}, these occlusion attributes can be combined to achieve a large number of more realistic occlusion types, which improves the occlusion diversity of large-scale face datasets. For the key points of the face, we obtain the 64 landmarks of the normal face by face alignment \cite{bulat2017how}. And at the same time we establish the key point mapping between the occluder and the face image, adjust the angle and size of the occluder, and finally make the occluder perfectly fit the normal person face image. This way, we construct a simulated occlusion face recognition dataset, covering 804,704 face images of 10,575 subjects.

\subsection{Statistics}
\label{ssec:subhead}
\begin{figure}[!t]
	\small
	\centering
	\vspace{-0.07cm}
	\hspace{-0.1cm}
	\includegraphics[width=0.9\linewidth]{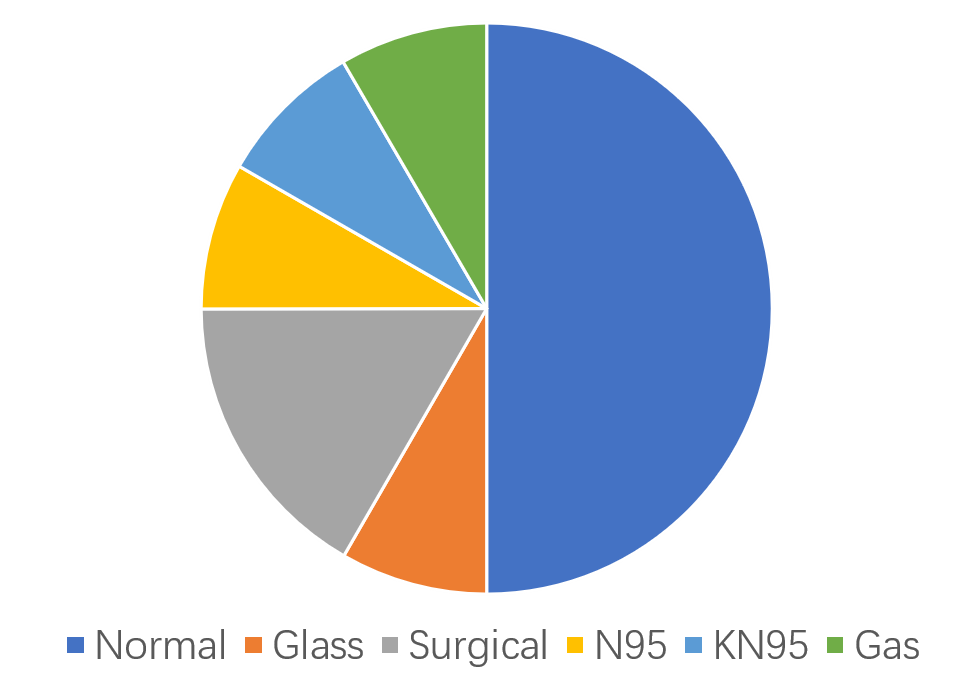}\\
	\vspace{-0.1cm}
	\caption{\small The distribution of the Webface-OCC dataset for various mask types.}
	\label{fig:04}
	\vspace{-0.3cm}
\end{figure}

After simulating all the 494,414 images, we keep part of the original face images to ensure the stability of the dataset in normal face image recognition. In practice, the simulated occlusion face dataset can be used along with their original unmasked counterparts. The dataset contains both occlusion and normal faces of the identities. The dataset is processed in terms of face alignment and image dimensions. Each image has a dimension of (112 $\times$ 112 $\times$ 3). The statistics of our simulated Webface-OCC is listed in Fig. \ref{fig:04}. The proportions of various masks and glasses are relatively uniform. It is worth noting that the surgical mask, as a very common mask occluder, gives more samples in our dataset. Note that the normal face images account for about half of the samples, as they will be more evenly distributed to each identity face image database. Thus each identity contains occluded and unoccluded face images. At the same time, we ensure that each face image can have effective recognition features, and avoid large-area occlusions that cause the image to be completely unrecognizable.

In view of the limitations of the occluder template, we improve the subjective diversity of occluders by adding various textures (about 30 types) in real life to the known template. Multiple occlusion types avoid the deep learning model from being too sensitive to the types of occluders.

\section{Experiments}
\label{sec:typestyle}
\subsection{Experimental Settings}
$\textbf{Datasets}$. In experiments, we use the large CASIA-Webface dataset \cite{yi2014learning} and our built Webface-OCC dataset for training and other face datasets (LFW \cite{huang2008labeled}, CFP-FP \cite{sengupta2016frontal}, AgeDB-30 \cite{moschoglou2017agedb}, as well as the recently proposed mask face recognition datasets (LFW-mask, CFP-FP, AgeDB-30-mask, RMFRD) \cite{wang2020masked} for testing. Webface is a large-scale face recognition dataset up to 10,000 subjects and 500,000 faces, thus suitable for model training. LFW is a public face verification benchmark dataset under unconstrained conditions.  CFP-FP contains 7000 images of 500 identities, each with 10 frontal and 4 non-frontal images. AgeDB-30 covers subjects of different ages. The  LFW-mask, CFP-FP-mask and AgeDB-30-mask dataset are the results of adding masks to the original datasets, and the data size and scale remain unchanged. RMFRD dataset contains 4015 face images of 426 people in the size of 250$\times$250 pixels, each with a normal face and several masked face images. The dataset is further organized into 7178 masked and non-masked sample pairs, including 3589 pairs of the same identity and 3589 pairs of different identities.

$\textbf{Benchmarks}$. To validate the effect of occlusion face recognition on the existing mask face recognition dataset and provide evaluation reference for researchers using the dataset, we retrain and compare six different CNN-based models on face recognition. The six baseline face recognition models include CenterFace \cite{wen2016a}, SphereFace \cite{liu2017sphereface}, FaceNet \cite{schroff2015facenet}, CosFace \cite{wang2018cosface} and ArcFace \cite{deng2018arcface}, along with a occlusion face recognition model MaskNet \cite{wan2017occlusion}. FaceNet and Arcface will be retrained into two versions using public WiderFace dataset \cite{yi2014learning} and our built dataset, respectively.

$\textbf{Evaluation Metrics}$. We evaluate the test models with precision and receiver operating characteristic curve (ROC) for face recognition.


\subsection{Implementation Details}
\begin{table*}[htbp]
	\vspace{-0.05cm}
	\normalsize
	\center
	\caption{\small Comparisons on face verification (\%) on LFW, CFP-FP, AgeDB-30 and RMFRD datasets. * denotes a retrained version.}
	\vspace{0.2cm}
	\footnotesize
	\setlength\tabcolsep{5pt}
	\centering
	\begin{tabular}{|c|ccc|cccc|}
		\hline
		Method & LFW \cite{huang2008labeled} & CFP-FP \cite{sengupta2016frontal} & AgeDB-30 \cite{moschoglou2017agedb} & LFW-mask \cite{wang2020masked} & CFP-FP-mask \cite{wang2020masked} & AgeDB-30-mask \cite{wang2020masked} & RMFRD \cite{wang2020masked} \\
		\hline
		CenterFace \cite{wen2016a}  &-  &-  &- &56.63  &56.35  &57.12 & - \\
		SphereFace \cite{liu2017sphereface}  &99.11  &94.38  &91.70 &58.22  &57.11  &57.02 & - \\
		FaceNet \cite{schroff2015facenet} &99.05  &94.12  &91.26 &59.68  &57.36  &57.89 & -\\
		CosFace \cite{wang2018cosface} &99.51  &95.44  &94.56 &60.70  &58.18  &58.14 & 61.06 \\
		ArcFace \cite{deng2018arcface} &99.53  &95.56  &95.15 &60.86  &58.04  &59.03 & 63.22\\
		\hline
		\hline
		MaskNet \cite{wan2017occlusion} &93.86  &80.56  &84.23 &83.22  &75.63  &72.62 & 68.78\\
		\hline
		\hline
		FaceNet(*) &97.98  &93.20  &90.19 &95.87  &85.63  &84.01 &77.56\\
		ArcFace(*) &99.01 &93.58  &93.27 &$\mathbf{97.08}$ &$\mathbf{87.18}$  &$\mathbf{86.07}$ & $\mathbf{78.25}$\\
		\hline
	\end{tabular}
	\label{tb:2}
\end{table*}

\begin{figure*}[!t]
	\small
	\centering
	\vspace{-0.07cm}
	\hspace{-0.1cm}
	\includegraphics[width=0.95\linewidth]{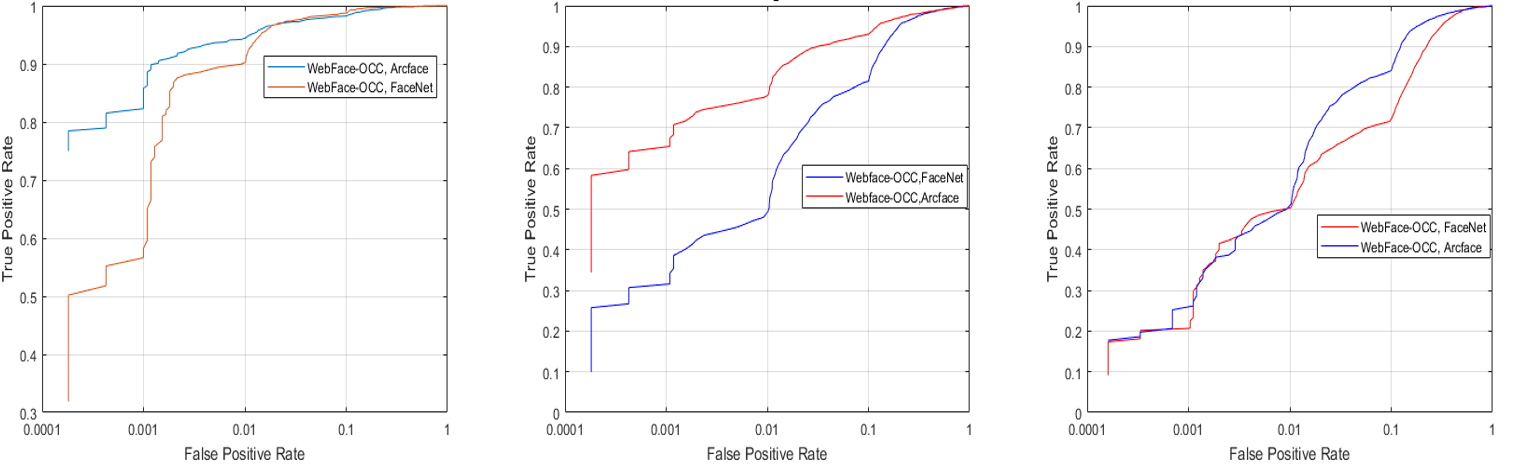}\\
	\vspace{-0.1cm}
	\caption{\small ROC curves of face verification on LFW-mask, AgeDB-30-mask and CFP-FP-mask datasets.}
	\label{fig:05}
	\vspace{-0.3cm}
\end{figure*}

For all face recognition models that need to be retrained, we employ the refined ResNet50 model proposed in ArcFace \cite{deng2018arcface} as our baseline CNN model. Our implementation is based on Pytorch deep learning framework, running on two NVIDIA 2080ti (12GB) GPUs. In training, the batch size is set to 128, and the training process is finished at 32K iterations. We extract the 512-dimension features for each normalized face in testing. To prevent over-fitting and improve the trained models' generalization, we perform data augmentation on the training set, like flipping.

The open-source RetinaFace \cite{deng2019retinaface} is used to detect occlusion faces from the raw images and obtain 68 facial landmarks. After performing similarity transformation accordingly, we align the face image and resize them to 112 $\times$ 112 pixels. 

\subsection{Results on Face Verification}
We evaluate the models trained on our dataset strictly following the standard protocol of unrestricted with labeled outside data \cite{huang2008labeled} and report the mean accuracy on test image pairs. 

As shown in Table \ref{tb:2}, * means that the model is retrained by our dataset. We divide the methods into two categories, including general face recognition and occlusion face recognition. As shown results, due to the influence of poses and ages, the accuracy on CFP-FP and AgeDB-30 is far lower than LFW. Obviously, the general models trained by Webface perform well on general face images. As a occlusion face recognition model, MaskNet has a significant reduction in the effect on the general face recognition dataset due to the additional occlusion elimination operations. It is worth noting that the model trained on our dataset still has an outstanding performance on the general face recognition dataset, which is only about 1\% less accurate relative to the original model. 

In comparison, the recognition accuracy in masked face recognition datasets has greatly shown the superiority of our dataset. In further examination, the retrained models still significantly outperform the original models (FaceNet and ArcFace). Specifically, relative to the original model ArcFace, the accuracy of retrained model on the four masked face recognition datasets has risen by 36.22\%, 29.14\%, 27.04\% and 15.03\%, respectively. This is really a remarkable gain in the face recognition task. Experiments show that our retrained model has a significant improvement in the performance of the occlusion face recognition dataset, without reducing the original effect on the general face recognition dataset.

Simultaneously, compared to the test on simulated masked face images, the recognition accuracy of all methods on the real masked face dataset is significantly reduced. Specifically, The large gap between the simulated masked faces and real ones can be attributed to the following facts. For the real-world masked faces, it is difficult to distinguish unknown occlusions accurately, and the dataset itself mostly comes from public figures, which deliberately disguise to avoid revealing identity.

To further verify the performance of our model, we then evaluate the results in the ROC indicator. The two retrained models are selected for display, where the results on LFW-mask, CFP-FP-mask and AgeDB-30-mask datasets are shown in Fig. \ref{fig:05}. Again, our retrained model is able to maintain a certain accuracy and stability when the FPR (false positive rate) is more than 1e-3. Admittedly, the accuracy drops sharply when the FPR is less than 1e-3, mainly due to the incomplete facial features. 

\section{Conclusion}
\label{sec:typestyle}

This research proposes a simulated occlusion face recognition dataset. We specially design an occlusion synthesis method, and apply it to the existing Webface dataset, thus obtaining a large number of occluded face images. Furthermore, we reasonably combine the original normal face image and the occluded face image to get our final dataset. Experimental results show that ArcFace retrained by our dataset can respectively give 97.08\% and 78.25\% accuracy on the simulation face datasets and the real-world masked face dataset, substantially outperforming the counterparts. In the future, we will further develop a universal occlusion face recognition algorithm on this basis.

\ifCLASSOPTIONcaptionsoff
  \newpage
\fi

{
	\bibliographystyle{IEEEtran}
	\bibliography{Trans}
}



%

\end{document}